\documentclass[conference]{IEEEtran}
\IEEEoverridecommandlockouts

\usepackage[T1]{fontenc}
\usepackage{float}
\usepackage{multirow}
\usepackage{array}
\usepackage{makecell}
\usepackage{enumerate}   
\usepackage[english]{babel}
\usepackage[utf8]{inputenc}
\usepackage{algorithm}
\usepackage{amsfonts}

\usepackage[noend]{algpseudocode}
\usepackage{optidef}
\usepackage{subcaption}
\usepackage[dvipsnames]{xcolor}
\usepackage{siunitx,booktabs}
\usepackage{xcolor}

\usepackage{amsthm}

\usepackage[justification=justified]{caption}

\usepackage{booktabs}
\usepackage{colortbl}
\usepackage{multirow}

\def\BibTeX{{\rm B\kern-.05em{\sc i\kern-.025em b}\kern-.08em
    T\kern-.1667em\lower.7ex\hbox{E}\kern-.125emX}}
\begin{document}

\title{Combinatorial Auctions and Graph Neural Networks for Local Energy Flexibility Markets}

\author{\IEEEauthorblockN{Awadelrahman M. A. Ahmed}
\IEEEauthorblockA{\textit{Department of Informatics} \\
\textit{University of Oslo}, Norway \\
aahmed@ifi.uio.no}
\and
\IEEEauthorblockN{Frank Eliassen,~\IEEEmembership{Member,~IEEE,}}
\IEEEauthorblockA{\textit{Department of Informatics} \\
	\textit{University of Oslo}, Norway \\
	frank@ifi.uio.no}
\and
\IEEEauthorblockN{Yan Zhang,~\IEEEmembership{Fellow,~IEEE}}
\IEEEauthorblockA{\textit{Department of Informatics} \\
	\textit{University of Oslo}, Norway \\
yanzhang@ieee.org}

}

\maketitle

\begin{abstract}

This paper \footnote{Accepted in The IEEE PES ISGT Europe 2023 (ISGT Europe 2023), Grenoble, France, on October, 2023.} proposes a new combinatorial auction framework for local energy flexibility markets, which addresses the issue of prosumers' inability to bundle multiple flexibility time intervals. To solve the underlying NP-complete winner determination problems, we present a simple yet powerful heterogeneous tri-partite graph representation and design graph neural network-based models. Our models achieve an average optimal value deviation of less than 5\% from an off-the-shelf optimization tool and show linear inference time complexity compared to the exponential complexity of the commercial solver. Contributions and results demonstrate the potential of using machine learning to efficiently allocate energy flexibility resources in local markets and solving optimization problems in general.

\end{abstract}


\section{Introduction}

The increasing adoption of photovoltaic (PV) energy at the distribution level due to the decreasing costs of solar and battery storage systems poses significant challenges for modern power systems. The European Commission has recognized the importance of managing local energy communities to integrate distributed energy resources and enable end-users to become active energy service providers \cite{european2017directive}. Energy flexibility refers to a system's capacity to utilize its resources in response to fluctuations in the net \cite{lannoye2012evaluation}. Prosumers who own PV systems can offer energy flexibility services at a local level, which can be combined and utilized in the energy market with the help of aggregators. In this market, prosumers serve as sellers, while the distribution system operator (DSO) is the buyer, and flexibility services serve as the commodity.

Combinatorial auctions allow owners of PV and energy storage systems (ESS) to bundle flexibility intervals, rather than bidding for them individually. For example, a prosumer with two PV production time intervals, $a$ and $b$, can choose to provide energy flexibility immediately in the same time interval, or store energy in the ESS and bid for a later use. Additionally, the prosumer can supplement any one of the PV production intervals with an off-peak interval, $c$, for example, at night, by using the ESS. Combinatorial auctions enable prosumers to submit more competitive bids by offering more choices and mitigating the risk of losing desirable flexibility-provision intervals. In this example, the prosumer can bundle all options as a list of mutually exclusive time-interval-combinations $ \big[ \{a\} , \{b\} , \{c\} , \{a \land b\} , \{a \land c\} , \{b \land c\} \big] $ with a corresponding list of power values. Combinatorial auctions have been successful in other domains, such as airport time slots allocation and spectrum auctions \cite{cramton2006combinatorial}, but they have not been thoroughly studied in the energy domain, especially in local energy markets. Previous research on local energy markets has focused on sequential auction mechanisms as in \cite{olivella2018local,xuincentive}, which do not satisfy the bundling need in local flexibility markets (LFMs). 

However, in LFMs, each time interval comprises multiple flexibility dividends that can be won by different bids and prosumers. This divisibility notion leads to complex winner determination problems (WDP) posing a challenge for implementing combinatorial auctions. This paper addresses these challenges by proposing a machine learning approach that efficiently solves the winner determination problem using graph neural networks (GNN). The approach considers multi-minded bidders proposing a reverse combinatorial auction framework for LFMs. By learning an end-to-end model that maps WDP instances to solutions, computation complexity is decoupled and efficient allocation time is achieved. This work mainly focuses on the management layer of energy market and does not address regulations or physical layer challenges.

We introduce the combinatorial auction framework in Section \ref{sec2} and discuss the LFM-WDP and its complexity in Section \ref{sec3}. Then, we propose our LFM neural combinatorial auction in Section \ref{sec4} and discuss our  GNN model for learning the LFM-WDP solution in Section \ref{sec5}. We present numeric results in Section \ref{sec6} and concluding remarks in Section \ref{sec7}.

\section{Local Flexibility Market Combinatorial Auction }\label{sec2}

\begin{figure}[htpb]
	\centering
	\begin{subfigure}[b]{0.5\textwidth}
		\includegraphics[scale=0.13]{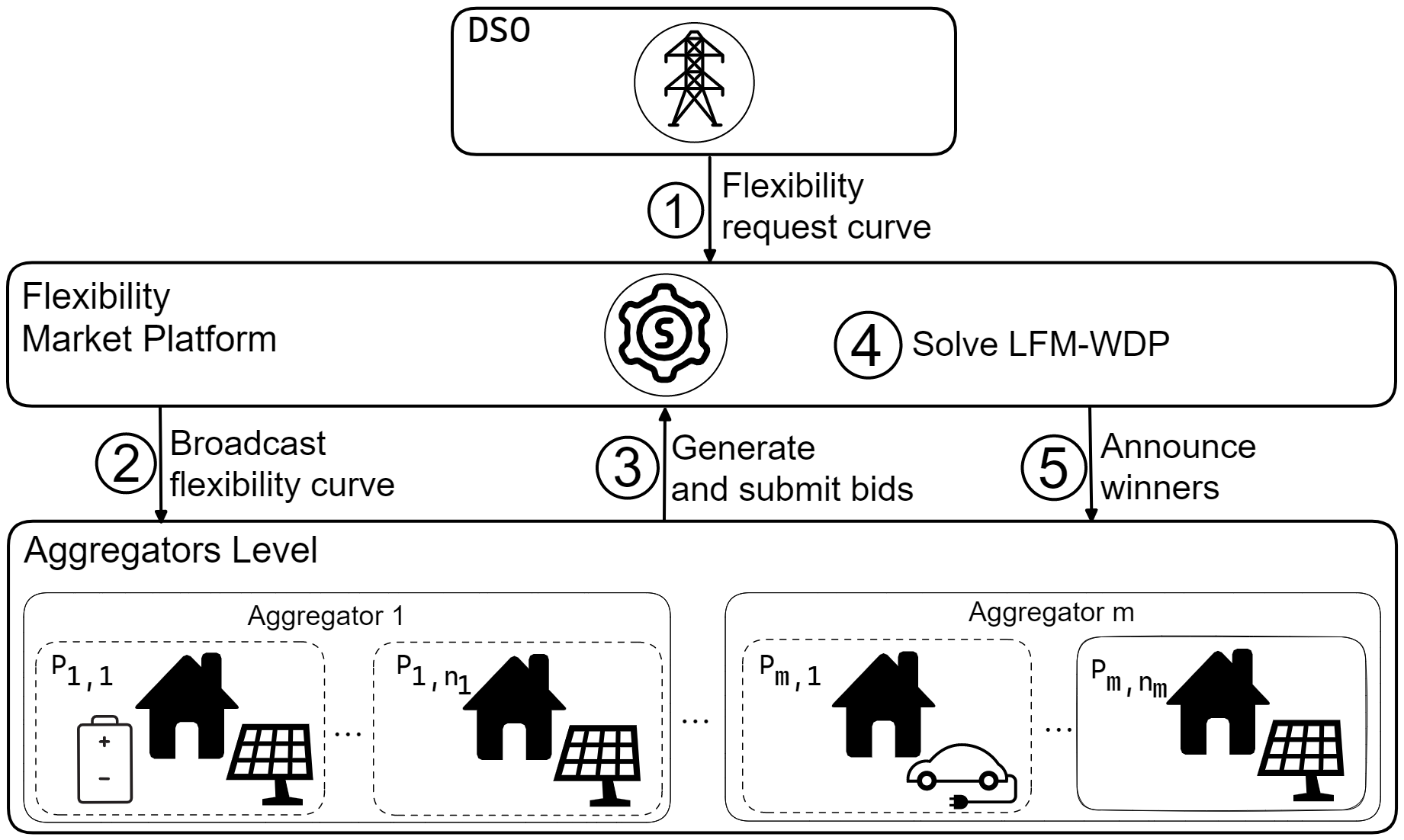}
		\centering
	\end{subfigure}
	\caption{LFM combinatorial auction temporal structure}\label{fsysmodel}
\end{figure}

We define the flexibility interval as the time period during which the DSO requires a provision of ramp-up (e.g., energy supply or load shedding) or ramp-down (e.g., energy reduction or load increase) in active power units. We assume that the DSO communicates its flexibility requirements through a flexibility curve representing the needed amount of flexibility units.  Figure \ref{fsysmodel} shows the proposed auction framework which involves the DSO as the buyer and aggregators as sellers, each managing a portfolio of prosumers with flexible resources and accessing their forecasting modules. Aggregators bid on behalf of their prosumers, considering preferences and available resources. The flexibility market operator evaluates bids and determines winners by solving an LFM-WDP. Next, we mathematically model the flexibility request and resources and define the bids' formats and auction objective.

\subsection{Flexibility modelling}\label{flexsection} 

\subsubsection{Flexibility curve}
The flexibility curve $ \mathcal{F}$ explicitly represents the amount of ramp-up or ramp-down requested by the DSO for each interval, with real power units as flexibility units. We represent $\mathcal{F}$ as $\{u_1,u_2,...,u_T\}$, where $ u_j \in \mathbb{Z}$ is the quantity of requested flexibility units at interval $j \in \mathit{I_M}=\{0,1,2,...,T\}$ and $ sign(u_j) $ indicates a ramp-up (+) or a ramp-down (-) flexibility request. An example of a DSO flexibility request when $T$ is 6 can be expressed as $\mathcal{F}=\{25,10,15,0,-30,-10\}$ megawatt, denoting requesting ramp-ups flexibility units at $ \{1,2,3\} $ and ramp-downs $ \{5,6\} $ flexibility units with no requirement at $ \{4\} $.

\subsubsection{PV system}

We investigate prosumers with fixed nominal PV production and no advanced peak-point-tracking technology, that have roof-top solar systems that follow a bell-curve pattern with peak power output at midday. In our study, we assume that PV systems are not connected to the grid during the bidding process, showing that they can only offer ramp-up services by contributing their production. We define $\mathit{I_{pv}}$ as the set of peak production intervals and $\sigma^{pv}_j$ as the forecasted PV power output in flexibility units in each interval $j$.

\subsubsection{PV-storage system}

As a practical consideration, we study the case where a portion of PV owners are equipped with sufficient energy storage systems (PV-ESS) which can be used to store energy either from their own PV units or the grid and can be used to feed energy to the grid. Under this consideration, PV-ESS can provide both ramp-up and ramp-down flexibility services. Hence, let $\mathit{I_{ss}} \subseteq \mathit{I_M}$ be the set of intervals of storage availability and $ \sigma^{ss}_j \leq u_j$ is the storage system charging/discharging power in flexibility units such that $ \sigma^{ss}_j \in \mathbb{Z} $ for each interval $ j $, where $ sign(\sigma^{ss}_j) $ indicates a ramp-up (+) or a ramp-down (-) flexibility service.


\subsubsection{Flexibility provision}
We can summarize the available flexibility provision options based on the resources of prosumers as follows. A PV system can provide a ramp-up service by connecting to the grid while an ESS can provide both ramp-up by discharging energy to the grid and ramp-down by charging energy from the grid. 

\subsection{Combinatorial Bid Format and Auction Objective}
In this study, we consider a multi-minded bidder scenario where bidders can submit multiple and mutually exclusive bids. This scenario differs from the single-minded bidder case where bidders are only allowed to to submit a single bid. Such mutual exclusivity nature is referred to as XOR bids in the standard scheme in combinatorial auctions logic \cite{cramton2006combinatorial}.


\subsubsection{Valuation function}

The flexibility request $ \mathcal{F}$ of $ \tau \leq T$ flexibility intervals forms a subset space of $2^\tau$ for possible bids. Hence, we define a valuation function to represent prosumer $p$ assigned value to the subset space, as $v_p:  2^\tau \rightarrow \mathbb{R}$. Now, let $\mathit{S}$ be a desired subset by prosumer $p$ which consists of $\{\sigma^{pv}_1,\sigma^{pv}_2,...,\sigma^{pv}_m\}$ produced by PV and $\{\sigma^{ss}_1,\sigma^{ss}_2,...,\sigma^{ss}_q\}$ fed from the ESS. The cost is a principal component of the valuation from the prosumer perspective. Therefore, we can model the valuation of subset $S$ as

\begin{equation}
	\label{costeq}
	v_p(S)= \sum_{m^{'}=1}^{m} \alpha_{m^{'}}.\sigma^{pv}_{m^{'}} + \sum_{q^{'}=1}^{q} \beta_{q^{'}}.\sigma^{ss}_{q^{'}} + \gamma_p(S)
\end{equation}
where $\alpha_{m^{'}}$ and $\beta_{q^{'}}$ are the cost coefficients of time interval $m^{'}$ delivered by the PV and time interval $q^{'}$ which is stored in the ESS, respectively. Note that $\alpha_{m^{'}}$ is ideally 0, as no cost is connected to the PV generation and $\beta_{q^{'}} =2(1-\eta)$ where $\eta$ is the charging and discharging efficiencies and $\gamma_p(S)$ is a term representing the prosumer profit for subset $S$.

\subsubsection{Bid format}
From a prosumer set $\mathit{N}=\{1,2,...,n\}$, prosumer $\mathit{p}$ can submit multiple bids and we denote the bid $\mathit{i}$ as $\mathit{b}{p_i}$ which offers flexibility quantities $\mathit{S}i$ from the flexibility curve $\mathcal{F}$. Then, we can calculate the prosumer's valuation $v_p(\mathit{S}{i})$ using (\ref{costeq}). The bid consists of the subset of flexibility items and the prosumer's value for that subset, denoted as ${\mathit{(S{p_i},v_p(S_{p_i}))}}$. The flexibility market platform receives bids from all prosumers through their aggregators as $\mathcal{B} ={ \mathit{(S_{1_1},v_1(S_{1_1})),... ,(S_{n_{\kappa_{n}}},v_n(S_{n_{\kappa_{n}}}))} }$, where $ \kappa_{p}$ is the number of bids submitted by prosumer $p$. The subset $\mathit{S}{p_i} = \{\sigma_{j}\}$ is a set of flexibility units in time intervals $ j\in\mathit{I_S}_{p_i} \subseteq \mathit{I_M}$, where $\sigma_j$ is the quantity of flexibility units. It is important to note that a prosumer may submit multiple bids, hence $\kappa_{p} \in \mathbb{Z^+}$.

\subsubsection{Auction objective}

The objective of the auction mechanism is to determine the optimal allocation set, which is a combination of bids that collectively provide the required flexibility represented by $\mathcal{F}$ with the minimum total cost. This is achieved by solving an NP-complete optimization problem, i.e., the LFM-WDP, while adhering to essential constraints. Details about LFM-WDP and solution is in the next sections.


\section{Local Flexibility Market Winner Determination Problem}\label{sec3}
Our LFM combinatorial auction aims to minimize the total cost while allocating flexibility intervals to prosumers based on their bids. In general, solving WDPs in combinatorial auctions is a challenging computation problem due to the NP-completeness of multidimensional weighted set packing problems, as proven in \cite{fujishima1999taming}. This challenge arises from the overlapping nature of bid items. Our contribution is to use machine learning to address this problem in the LFM-WDP.
%

\subsection{ LFM-WDP Formulation} 
We mathematically formulate the LFM-WDP as a combinatorial optimization problem considering the divisibility nature of flexibility items mentioned earlier. Given a flexibility curve $\mathcal{F}=\{u_j\}$, where $j\in\mathit{I_M}$ and a bid space $\mathcal{B} ={\mathit{(S_{p_i},v_p(S_{p_i}))}}$, $ \forall i\in {1,..., \kappa_{p}}$ and $\forall p\in N$, the objective is to find the set $\mathcal{A}$ of winning bids that minimizes total costs. We formulate the LFM-WDP as 


\begin{mini!}[2]
	{\mathbf{x}}{J= \sum_{ \forall i,\forall p}   v_p(\mathit{S}_{p_i}) x_{p_i}} {\label{WDP0}}{}\label{WDPa}
	\addConstraint{\sum_{\forall i,\forall p} \sigma_{j}(\mathit{S}_{p_i}) x_{p_i} }{\geq u_j; \ \forall  j \in \mathit{I_{S_{p_i}}} } \label{WDPc1}
	\addConstraint{\sum_{\forall i}x_{p_i}}{\leq 1; \ \forall  p \in \mathit{N} } \label{WDPc2} 
	\addConstraint{x_{p_i}}{\in \{0,1\}; \ \forall i,\forall p }  \label{WDPc4}
\end{mini!}  
where $ \mathbf{x} $ is the decision variables' vector. The solution should satisfy the set of constraints we discuss below.

Firstly, it is plausible that the bid space may not exactly match the required flexibility curve due to the limitations in prosumers' bids controlled by their PV and ESS nominal capacity. As a result, the constraint $\sum \mathcal{A} = \mathcal{F}$ may not be feasible. To account for this, we assume the DSO can tolerate allocating more bids than needed to exceed flexibility requirements, rather than allocating less flexibility for lower costs. This is shown in the relaxation $\sum \mathcal{A} \geq \mathcal{F}$ and encourages prosumer engagement, increasing the \textit{thickness} of the long-term market. Constraint (\ref{WDPc1}) satisfies this requirement, and the relaxation is implicitly upper-bounded by the minimization problem (\ref{WDPa}). Secondly, prosumers are constrained to submit mutually exclusive bids, meaning that they can win at most one bid. This constraint is expressed in (\ref{WDPc2}). Thirdly, we impose an integrality constraint (\ref{WDPc4}) to account for the indivisibility of bids, as prosumers are unwilling to accept partial bids. Note that \textit{bid} indivisibility is distinct from \textit{item} divisibility in the flexibility curve. The latter allows multiple units to be allocated to different prosumers within a given time interval, whereas the former dictates that each bid is a unitary entity that can only either be won or lost.

\subsection{LFM-WDP Complexity}\label{complex} 


The complexity of an optimization problem is typically measured by its encoding length, which is the number of binary symbols required to store an instance of the problem. Tractable problems have a number of operations that are bounded by a polynomial function in the encoding length. For our LFM-WDP, the number of bids made by  $ n $ prosumers is in $ \mathcal{O}(n \times 2^{c . T }) $, which is exponential in the number of time intervals and biddable items. Although our LFM-WDP in (\ref{WDP0}) is an integer linear programming problem, it has two characteristics that place it in the difficult category. 

Firstly, the integrality condition in (\ref{WDPc4}) makes the feasible region non-convex and shifts the problem to mixed-integer linear NP-hard problems. Secondly, the correlation between valuations $v_p$ and their corresponding multiplicities ($ \sum_{\forall j \in \mathit{I_{S_{p_i}}} } \sigma_{j}$) in each bid $\mathit{b}_{p_i}$ determines the problem's classification as either uncorrelated or strongly-correlated, with the latter being more challenging. Our LFM-WDP belongs to the \textit{strongly-correlated} problem class. Although the correlation between the values and multiplicities of items appears to be stochastic, our problem exhibits significant correlation. For example, when instances are generated with equation (\ref{costeq}) for up to 24 intervals in the flexibility curve and 200 submitted bids, the correlation factor is approximately 0.9. This class of problems has been extensively analyzed in the Knapsack problem literature \cite{pisinger2005hard} and to address this difficulty, we leverage machine learning.

\section{LFM Neural combinatorial auction}\label{sec4}
\begin{figure*}[htpb]
	\centering
	\captionsetup{width=0.75\linewidth,format=hang}
	\begin{subfigure}[b]{\textwidth}
		\includegraphics[width=0.70\linewidth]{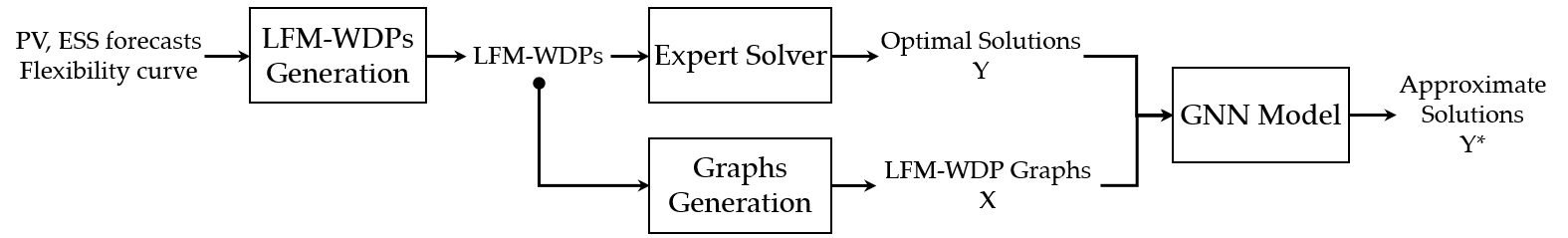}
		\centering
	\end{subfigure}
	\caption{LFM neural combinatorial auction design framework}\label{blockdiag}
\end{figure*}

Despite previous discussions on the complexity and difficulty of LFM-WDP, we intend to utilize the inherent similarity among the problem instances. We hypothesize that LFM-WDP instances follow an underlying unknown probability distribution  $ \mathcal{P} $, which is determined by the nature of LFMs. This research aims to use graph neural networks (GNNs) as a machine learning framework to learn this unknown probability distribution. The primary objective is to develop a machine learning model capable of mapping LFM-WDP instances to their optimal solutions. This will be achieved by adopting a supervised learning approach, imitating an off-the-shelf solver as an expert system and producing plausible optimal solutions. Our specific aim is to learn a function $h: \mathcal{X} \rightarrow \mathcal{Y}$, where $ \mathcal{X} $ represents the LFM-WDP representation, and $ \mathcal{Y} $ denotes the corresponding target set of solutions. To achieve this, we followed our design framework shown in Figure \ref{blockdiag} .


\subsection{LFM-WDP Instance Generation and Expert Solver}\label{instgen}
One challenge of our machine learning approach is the lack of gold standard data, i.e., $ \mathcal{X} \times \mathcal{Y} $ pairs. To address this, we have developed a procedure for generating LFM-WDPs. We adopt a bottom-up approach using a real PV production dataset from prosumers \cite{ausgridds}. Specifically, we start by forming a set of possible bids that prosumers would bid for and then generate the corresponding flexibility curve. This approach provides more control over the complexity of the generated instances compared to randomly generating flexibility curves and then generating bids.


From a prosumer perspective, a biddable time interval is one that has PV production $ u_{prod} $ greater than a threshold portion $\epsilon$ of its full capacity $u_{max}$. Then, we define prosumer $ i $'s biddable set as	$I_{pv}= \big\{ j : u_{prod_j} \geq \epsilon \cdot u_{max_j}; \ \ j \in I_M\big\}$ with its PV production set as $ U_{pv}= \big\{ u_{prod_j}; \ \ \forall j \in I_{pv}\big\}$.


Considering the two types of prosumers, those who own only PVs and those equipped with ESS, the former can only bid on all subsets of time intervals in $I_{pv}$ with their corresponding values from $U_{pv}$ to provide ramp-up services, resulting in $\sum_{k=1}^{|I_{pv}|}{{|I_{pv}|}\choose {k}}$ bids, the latter can bid on additional ramp-up and ramp-down intervals.

We define the availability of ESS time intervals as any time interval when the PV production is not sufficient as $ I_{ss}= \big\{ j : u_{prod_j} < \epsilon.u_{max_j}; \ \ \forall j \in I_M \big\}$. Then the biddable intervals for this group of prosumers is as  $ I_{pv\_ss}  =  \big\{ \{ j : u_{prod_j} \geq \epsilon.u_{max_j}\} \cup \{ k : u_{prod_k} < \epsilon.u_{max_k}\} ; \forall j, k \in I_M \big\}$ thus the bids space that serves for the ramp-up flexibility requests contains all subsets of $ S_{pv\_ss} \subseteq \big\{ I_{pv\_ss}: \big( \big|I_{pv\_ss}:j \in I_{pv} \big| + \big|I_{pv\_ss}:k \in I_{ss} \big|\big) \leq \big|I_{pv}\big|\big\} $.





With ESS, prosumers can bid on ramp-down intervals if their ESS is fully discharged, for every subset $S \in I_{pv_ess}$, with up to $|S|$ ramp-down intervals. We calculate the price using formula ($\ref{costeq}$) and accumulate the flexibility curve proportionally to the sum of the bids, as  $\mathcal{F} = \bigg\{\eta \sum\limits_{i=1}^{\kappa}  \sigma_j(\mathit{S}_i); \ \forall j \in I_M\bigg\}$ where $ 0 <\eta \leq 1$ is the proportionality factor and we use it to control the correlation and hence the problems complexity, $\kappa$ is the number of bids, $\sigma_j(S_i)$ is the offered flexibility by the subset $ S_i $ for time interval $ j $.

%


Commercial mixed-integer linear programming (MILP) solvers use advanced algorithms to converge to optimal solutions, but they still make heuristic decisions during runtime. For example, in branch-and-bound algorithms, variable and node selections are critical decisions, while the Gomory cut in cutting-plane approaches requires computation time. We aim to learn these underlying heuristics during the GNN-based model's training phase.

To obtain the corresponding set of labels $\mathcal{Y}$, we represent our LFM-WDPs as equations set (\ref{WDP0}) and use a mixed-integer linear programming solver (MILP) as an expert based on the data generated in the previous step.

%

\subsection{LFM-WDP Graph Representation}

To effectively utilize GNNs for solving the LFM-WDPs, it is critical to create an appropriate graph representation. A graph is a mathematical structure that consists of nodes and edges, representing measurable elements and their relationships. Nodes and edges have quantifiable features, which enable the analysis and comparison of different graphs. The proposed graph representation for LFM-WDPs includes bid nodes $\textbf{x}$, flexibility (Flex) nodes $\beta$, and mutual exclusiveness (MUX) nodes $\alpha$. The graph representation addresses the similarity issue of node features by incorporating neighbouring node features into the analysis.

Each bid node $x^i_p$ corresponds to a bid $i$ from prosumer $p$, with a feature vector [$f^i_p$] concatenating the bid price and items, i.e., $ [f^i_p] = [v_p(S_i) \mathbin\Vert S_i] $. Flexibility nodes $\beta_\tau$ represent the total capacity of each interval in the flexibility curve $\mathcal{F}$, with a distinctive feature vector $u_\tau \in \mathbb{R}$. MUX nodes $\alpha_i$ represent the XOR condition of prosumer $i$ bids, with features set to unit vectors.

Edges between nodes capture the relational structure of the problem. In this graph representation, nodes of the same type are not connected, and it is categorized as a tri-partite graph. Bid nodes are connected to the corresponding flexibility nodes by undirected edges, with bid quantity [$\sigma_{i\tau}$] as the edge feature. Prosumer nodes are connected to the corresponding MUX nodes by undirected edges, with features set to unit vectors.

\begin{figure}[htpb]
\centering
\begin{subfigure}[b]{0.5\textwidth}
\includegraphics[width=0.85\linewidth]{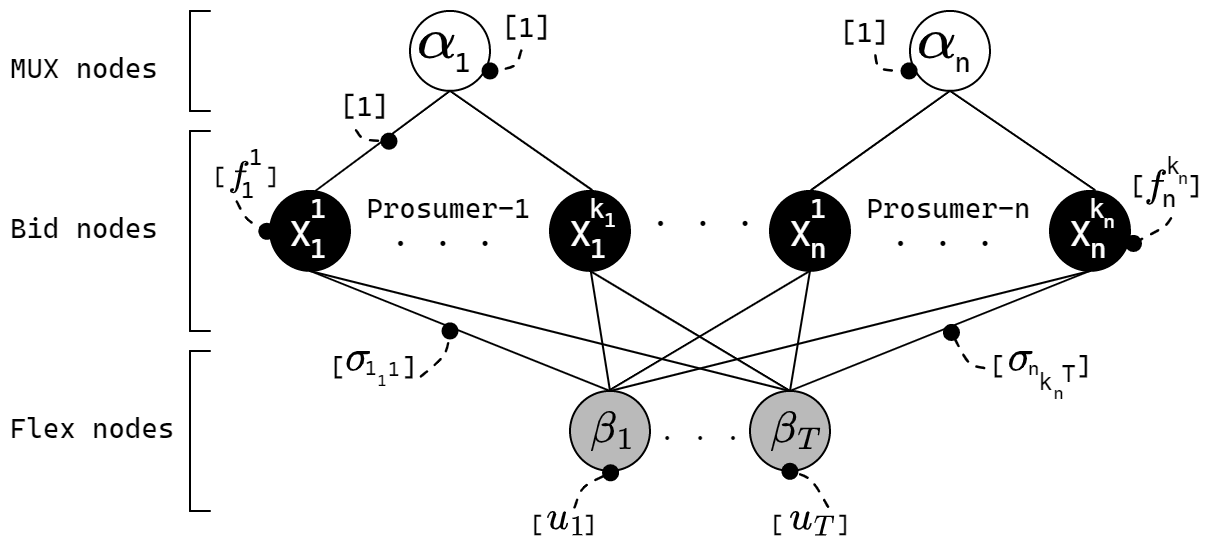}
\centering
\end{subfigure}
\caption{LFM-WDP graph representation showing the three types of nodes with edges between them}\label{wdpgraph}
\end{figure}

Thus the proposed graph consists of bid nodes $\mathcal{\text{X}}$, flexibility nodes $ \boldsymbol{\beta} $, mutual exclusiveness nodes $ \boldsymbol{\alpha} $, and edges $\mathcal{E}$ is denoted by $\mathcal{G(\text{X}, \boldsymbol{\beta}, \boldsymbol{\alpha},E)}$. Its adjacency matrix, $\textbf{A} \in \mathbb{R}^{n_{nodes} \times n_{nodes}}$, with $n_{nodes}=\kappa + T + n$, serves as a measure of the LFM-WDP size.

\section{GNN Model for the Neural Solver}\label{sec5}

We tackle the LFM-WDP as a multi-label node classification problem in which we aim to learn a function $h \colon \mathcal{G} \to \mathcal{Y} \in \mathbb{Z^{\kappa}}$ given $\mathcal{G(\text{X}, \boldsymbol{\beta}, \boldsymbol{\alpha},E)}$. Unlike conventional classification tasks with mutually exclusive class labels, in our case,  multiple nodes can be assigned to the same class. Our approach is fully-supervised, where a model is trained to classify nodes across multiple labeled graphs. We propose a combinatorial auction graph neural network and discuss its design choices, including inter-node communication, message transformation and aggregation, network depth, learning objective and loss function. This approach differs from semi-supervised node classification problems, which focus on assigning labels to partially labeled nodes in a single massive graph.

\subsection{Computation Graphs and Messages} 

\begin{figure}[htpb]
	\centering
	%
	\begin{subfigure}[b]{0.5\textwidth}
		\includegraphics[width=0.85\linewidth]{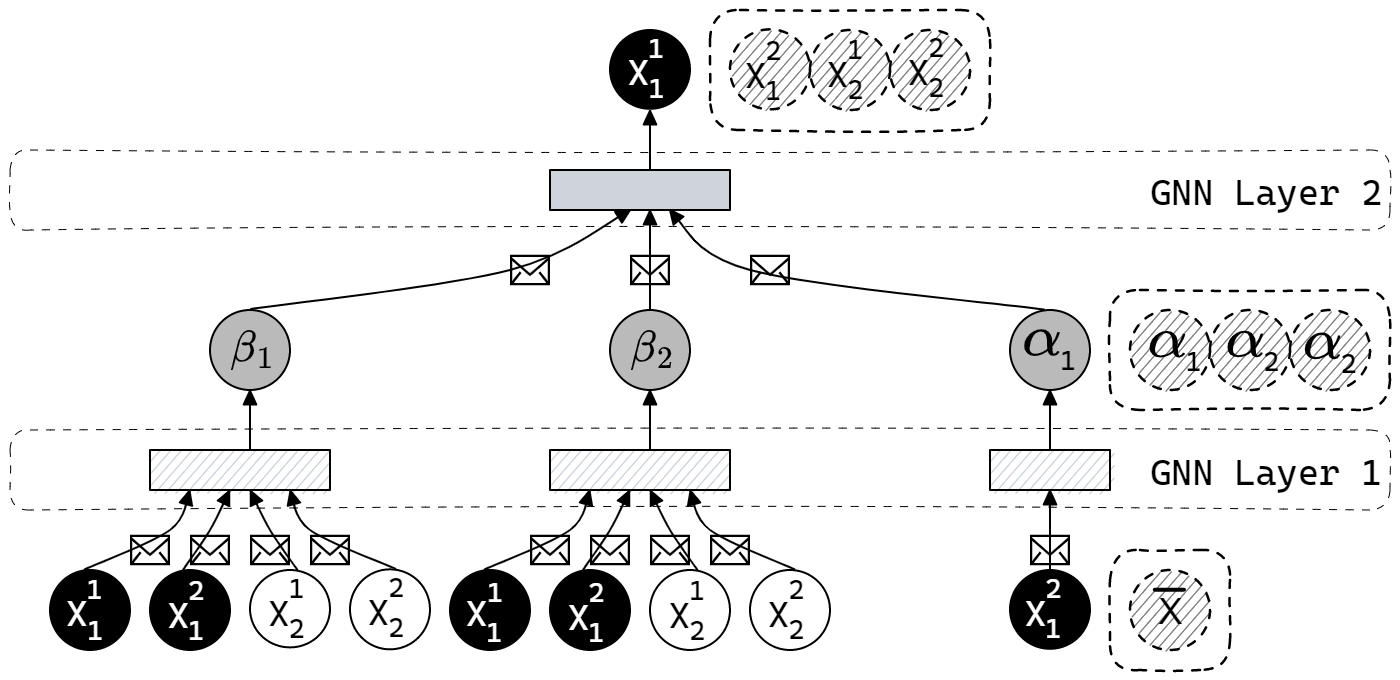}
		\centering
	\end{subfigure}
	\caption{Computation graph of two hops. To calculate the features of layer 2 output, we involve the corresponding  $\alpha$ from layer 1 output and the complementing x from the input. }\label{redgraph}
\end{figure}

To demonstrate the computation of each node's features considering its neighborhood, we unfold our proposed graph representation to obtain a computation graph for each node. A simplified example of an LFM-WDP of 2 prosumers, each one submits 2 bids competing on 2 flexibility intervals is illustrated in Figure \ref{redgraph}. Each node aggregates information from its neighbors to compute its feature representation. Note that our proposed heterogeneous tri-partite graph representation offers a simple yet expressive advantage, that is bid nodes communicate with their constraint-type neighbors through a two-hop computation. The immediate neighbors of a bid node are only constraint-type nodes, either $\beta$ or $\alpha$, and vice versa. Therefore, a two-hop depth is sufficient to capture the overall graph structure, as bid nodes can only communicate their information through constraint nodes. This simplifies our GNN model, as each bid node's state is updated after being aware of the constraints' state.


We use a message function $\mathcal{M}(r)$ to convert the feature representation of a neighboring node $r$ to a message $m_r$. This message is sent to node $q$, which has its own feature representation $h_q$ and set of neighboring nodes $r_q$. We also generate a self-message from node $q$'s features and incorporate it into the aggregation process using an order-invariant function $AGG$ and non-linearity function $\Gamma$. This process generates the message for step $t+1$ which computed as: 

%
\begin{equation}
\label{message}
h^{t+1}_q= \Gamma \bigg( \Big< \text{AGG} \big\{\mathcal{M}(h^{t}_r), \forall r \in r_q \big\}, \mathcal{M}(h^{t}_q) \Big> \bigg)
\end{equation}
The intuition behind this is that nodes in the graph seek a condensed representation of their neighbors' state, which is merged with their own state to form a new state.

%
%



\subsection{GNN Design} 
To update their states, nodes need to decide how many hops to use when aggregating messages from their neighbors to update their state. In a cyclic graph, this can lead to over-smoothing and saturation at the majority class. We avoid this issue by leveraging the unique connectivity of bids in our graph. Bids competing for the same flexibility interval are connected through representative flexibility nodes $\beta$, while bids from the same prosumer are linked through mutual exclusiveness nodes $\alpha$. This design ensures that two hops capture node dependencies and prevent over-smoothing.

%


We wrap our GNN model within a classifier function to classify nodes in the LFM-WDP problem. To address the targets imbalance, we introduced a weighting factor $\lambda$ in the binary cross-entropy loss $\ell$, which is the proportion of the minority class in the data. The loss function is defined as:

\begin{equation}
	\label{loss1}
	\ell = \frac{-1}{\kappa}\sum_{i=1}^{\kappa}\lambda_i \Big[ y_i \log(p(y_i)) + (1 - y_i)\log(1-p(y_i))\Big]
\end{equation}
where $\lambda_i$ is the weight of the class of node $i$, and $y_i$ is the true label of node $i$. This loss term calculates the difference between the bids allocations generated by the GNN-based model and those generated by the expert solver.

Additionally, we included the mean square error between expert solver optimal value $J$ and model optimal value $J^*$ calculated from predicted optimal allocations in the loss function to optimize the learning process towards expert solver optimal solutions. The weighting factor $\zeta$ balances the contribution of this term to the total loss. Then the total loss function be

\begin{equation}
	\label{loss}
	\mathcal{L} = \frac{1}{Q} \sum_{i=1}^{Q} \ell_i + \zeta . (J_i-J^*_i)^2
\end{equation}
where $\ell_i$ is the loss of the $i$-th problem instance and $J_i$ and $J_i^*$ are the expert solver optimal value and model optimal value, respectively, of the $i$-th problem instance. The total loss is minimized during training on $Q$ problem instances.

Furthermore, in order to ensure compliance with the XOR constraint, we assign the bid with the highest probability value from the classifier within the conflicting set if conflicts arise.

\section{Numeric Evaluation}\label{sec6}
\begin{figure}[h]
	\subcaptionbox{F1 score of GNN for different LFM-WDP complexities compared to a reference FNN \label{fig1:a}}{\includegraphics[width=1\linewidth,height=49mm  ]{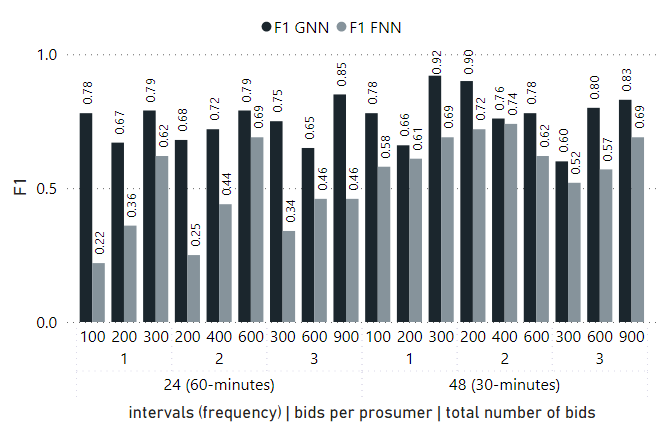}}\vfill 
	\subcaptionbox{Deviation percentage of the GNN's optimal value from the expert solver and the normalized root mean square difference (NRMSD\%) \label{fig1:b}}{\includegraphics[width=1\linewidth,height=49mm]{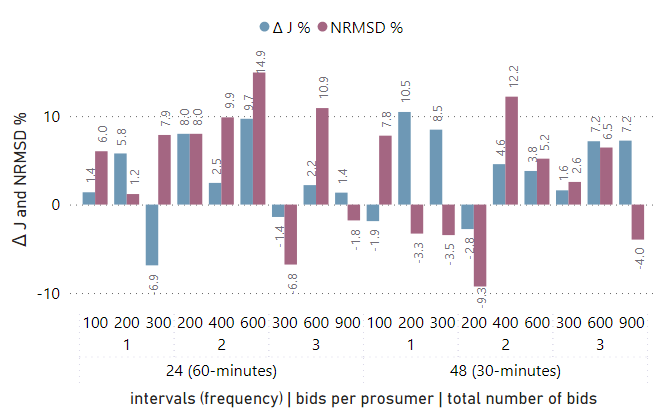}}\vfill
	\subcaptionbox{Solver exponential computation time VS LFM-WDP complexity  \label{fig2:a}}{\includegraphics[width=1\linewidth,height=33mm]{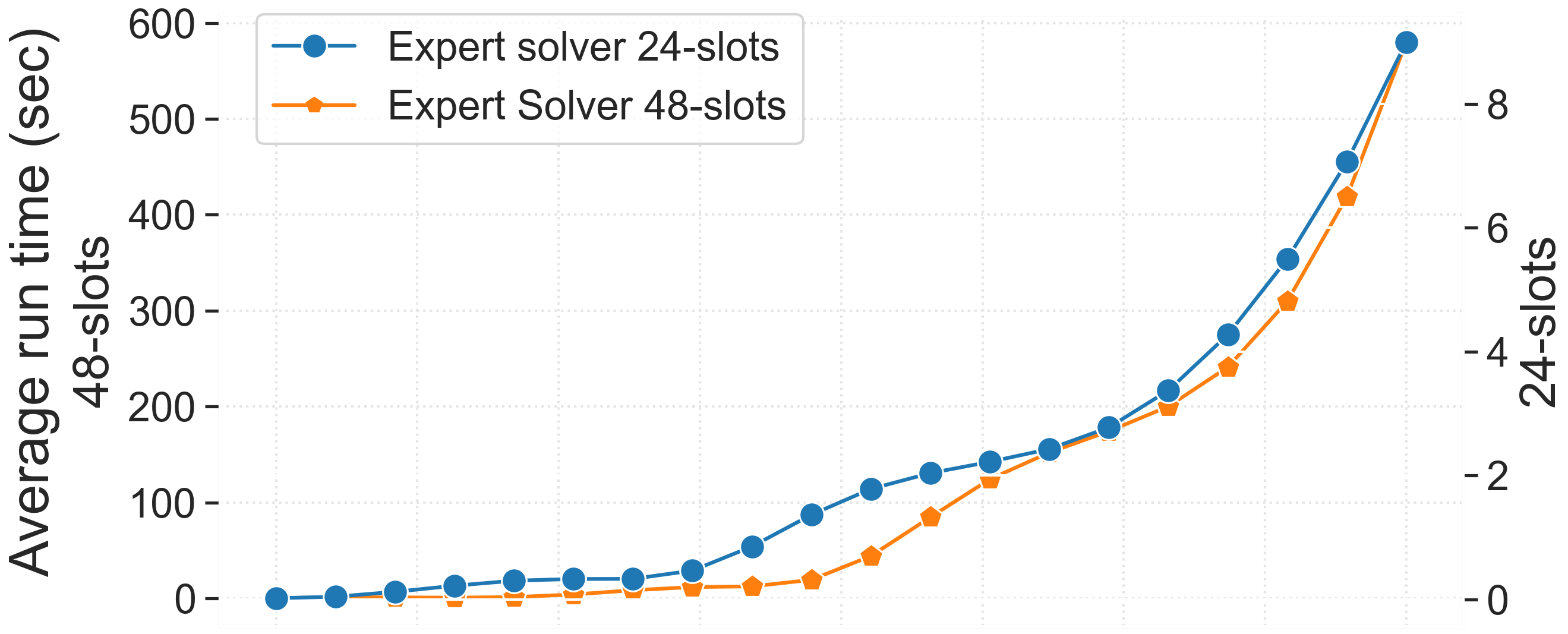}}\vfill%
	\subcaptionbox{GNN linear inference time VS LFM-WDP complexity  \label{fig2:b}}{\includegraphics[width=1\linewidth,height=35mm]{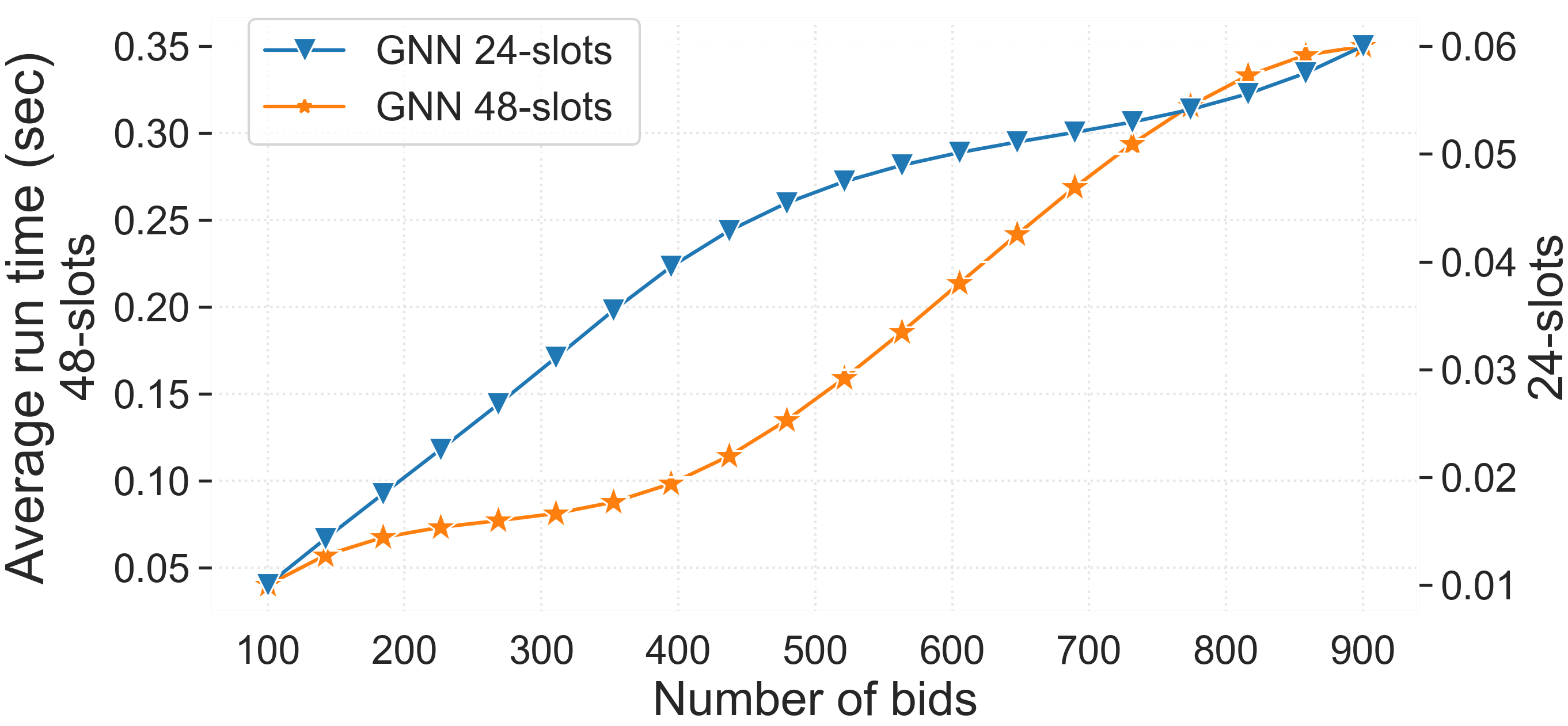}} 
\end{figure}

We quantitatively evaluate and analyze our approach using the solar home electricity dataset \cite{ausgridds}, which provides 30-minute which we resampled to 60-minute measurements of 300 homes' rooftop solar systems. We create LFM-WDP instances and their graph representations using the bottom-up generation procedure described in section \ref{instgen}, and use the Gurobi solver \cite{gurobi} as an expert to generate target solutions. We vary the number [100, 200, 300] of homes and bids per prosumer [1,2,3] to analyze model sensitivity, resulting in 18 cases in Figure \ref{fig1:a}. We produce a graph representation for each problem instance based on specific parameters that are the number of prosumers, the number of bids per prosumer, the prices, units of each bid and the DSO flexibility curve. We use the PyG library \cite{Fey2019} for graph generation, and the model consists of a 2-layer feed-forward network with ReLU activation for state update, and a 2-layer classifier appended to the GNN output. Each case is split into training and test sets and trained with a learning rate of 0.0001.

To ensure a thorough evaluation, we have developed four evaluation metrics. Firstly, we evaluate our approach's ability to produce optimal allocations that match those produced by the expert solver. Due to imbalanced data, we utilize the macro F1-score metric, which assesses the metric for each class independently and calculates the average and we compare it with a reference feed-forward neural network (FNN) base model as in Figures \ref{fig1:a}. The GNN achieved 0.76 F1 average score. Secondly, we calculate the deviation percentage of the GNN's optimal value from that of the expert solver's reference optimal value, denoted as $ \Delta J $. Thirdly, while the F1-score and $ \Delta J $ quantify the GNN's closeness to the expert solver's solution, we moreover evaluate how well the allocations fulfill the DSO request $\mathcal{F}$. For this, we introduce the normalized root mean square difference (NRMSD), measured as


\begin{equation}\label{nrmseexpert}
	\text{NRMSD \ } = \frac{1}{T} \sum_{t=1}^{T}  \frac{ \sum_{k=1}^{\kappa} \mathcal{A}_k - \mathcal{F} }{\mathcal{F}}
\end{equation}
we  then calculate the percentage difference between the GNN and the expert solver NRMSD values and denote it as $\Delta \text{ NMRSD}$. As in Figure \ref{fig1:b}, the average $\Delta \text{NMRSD}$ is 6.75\%, which indicates that the model allocations exceed the DSO flexibility request by an average of 6.75\%. This results leads to an average $\Delta J$ of 4.52\%. 

Lastly, as this work aims to use machine learning to approximate the bids' allocation process, reducing the complexity of the LFM-WDP computation. We found that the GNN model is 98\% more computationally efficient than our expert solver. However, there is an off-line training time overhead averaging 30 minutes. Figures \ref{fig2:a} and \ref{fig2:b} demonstrate the exponential relationship between the expert solver computation time and the number of bids, with an exponential increase on the y-axis scale.

\section{Conclusion}\label{sec7}

This paper proposes a new combinatorial auction framework for local energy flexibility markets leveraging graph neural networks to address the winner determination problem. Through experimentation on various problem complexities, the proposed models demonstrate a linear time complexity compared to the expert solver's exponential time complexity. The results also indicate that, on average, the proposed models achieve a 76\% match with expert solver allocations, while maintaining an average deviation of less than 5\% from the optimal value. Furthermore, the average deviation in meeting the flexibility needs of the system operator using the proposed models is found to be below 7\% compared to the expert solver.

\bibliographystyle{IEEEtran}

\bibliography{LFM_proposal3}


\end{document}